\documentclass[11pt,letterpaper]{article}
\usepackage{emnlp2017}
\usepackage{times}
\usepackage{latexsym}
\usepackage{latexsym}
\usepackage{multirow}
\usepackage{arydshln}
\usepackage{amsmath}
\usepackage{rotating}
\usepackage{color}
\usepackage[small]{caption}
\usepackage{subcaption}

\usepackage{adjustbox}

\emnlpfinalcopy

\def\emnlppaperid{5}

\newcommand{\ptheta}{p_{{\boldsymbol \theta}}}

\title{Unlabeled Data for Morphological Generation With Character-Based Sequence-to-Sequence Models}

\author{Katharina Kann \and Hinrich Sch\"{u}tze \\ 
LMU Munich, Germany \\
kann@cis.lmu.de}

\date{}

\newcounter{notecounter}
\newcommand{\enotesoff}{\long\gdef\enote##1##2{}}
\newcommand{\enoteson}{\long\gdef\enote##1##2{{
\stepcounter{notecounter}
\large\bf
\hspace{1cm}\arabic{notecounter} $<<<$ ##1: ##2
$>>>$\hspace{1cm}}}}

\usepackage{color}
\usepackage{xcolor}
\usepackage[disable]{todonotes}   
\newcommand{\note}[4][]{\todo[author=#2,color=#3,size=\scriptsize,fancyline,caption={},#1]{#4}} 
\newcommand{\katha}[2][]{\note[#1]{Katharina}{yellow!40}{#2}}   
   
\enoteson
\enotesoff

\begin{document}

\maketitle

\begin{abstract}
We present a semi-supervised way of training a character-based encoder-decoder recurrent neural network 
for morphological reinflection, the task of generating one inflected word form from another.
This is achieved by using unlabeled tokens or random strings as training data for an autoencoding 
task, adapting a network for morphological reinflection, and performing multi-task training.
We thus use limited labeled data more effectively, obtaining up to $9.9\%$
improvement over state-of-the-art baselines for 8 different languages. 
\end{abstract}

\section{Introduction}
\label{intro}
\enote{kk}{Find some citation for morphologically rich languages.}
Morphologically rich languages use inflection---the adaptation of a surface form to its syntactic
context---to mark the properties of a word, e.g., \textit{gender} or \textit{number} of nouns or \textit{tense} of verbs.
This drastically increases the type-token ratio, and thus negatively effects 
natural language processing (NLP), making morphological analysis and generation an important field of research.

In this work, we focus on morphological reinflection (MRI), the task of mapping one inflected form of a lemma to another, given the morphological 
properties of the target, e.g., (\textit{smiling}, \textit{PastPart}) $\rightarrow$ \textit{smiled}. The lemma 
does not have to be known. Recently, there have been some advances on the topic, motivated by the SIGMORPHON 2016 shared task on morphological
reinflection \cite{cotterell-et-al-2016-shared}
and the CoNLL-SIGMORPHON 2017 shared task on universal morphological reinflection \cite{cotterell-conll-sigmorphon2017}. 
In 2016, neural sequence-to-sequence models, specifically attention-based encoder-decoder models,
outperformed all other approaches by a wide margin \cite{faruqui-EtAl:2016:N16-1,kann2016med}.
However, those models require a lot of training data, while in contrast many morphologically rich languages are low-resource,
and little work has been done so far on neural models for morphology in settings with limited training data.
This makes sequence-to-sequence models
not applicable to morphological generation in most languages.
\begin{figure}[t]
  \centering
  \includegraphics[width=.93\columnwidth]{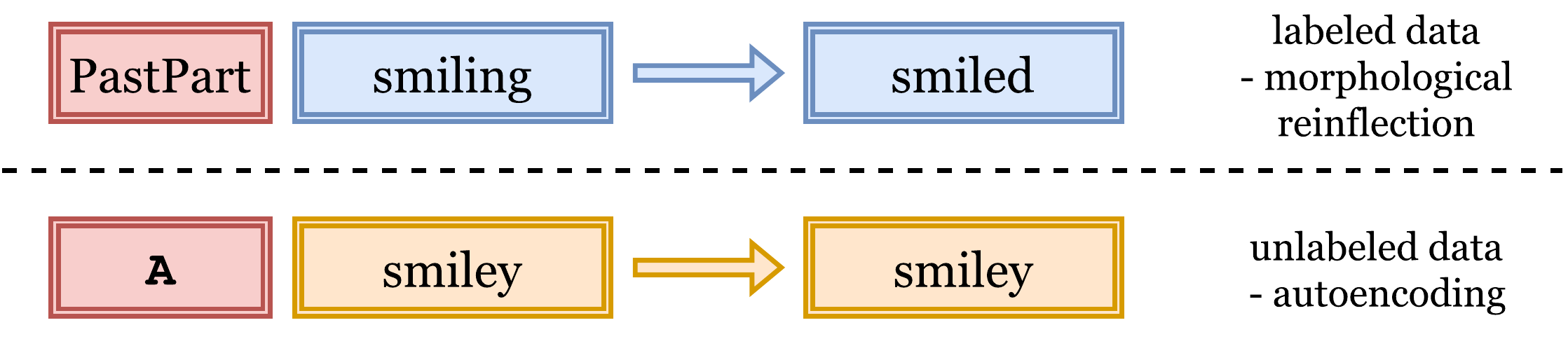}
  \caption{Examples for labeled and unlabeled input. The content of the red boxes (very left in both rows) signalizes if the sample belongs to the 
  MRI task or the autoencoding task.}
  \label{fig:example}
  \vspace{-.5cm}
\end{figure}

An abundance of \textit{unlabeled} data, in contrast, can be assumed available for each language in the focus of NLP. 
Thus, we propose a semi-supervised training method for a state-of-the-art encoder-decoder network for MRI using both
labeled and unlabeled data, mitigating the need for time-expensive annotations.
We achieve this by treating unlabeled words as training examples for an \textit{autoencoding} \cite{vincent2010stacked} task and multi-task training (cf. Figure \ref{fig:example}).
We intuit the following reasons why this should be beneficial:
(i) The decoder's character language model
can be trained using unlabeled data.
(ii) 
Training on a second task reduces the problem of overfitting.
(iii) By forcing the model to additionally learn autoencoding, we give it a strong prior to copy the input string. This
might be advantageous as often many forms of a paradigm share the same stem, e.g., \textit{\textbf{smil}ing} and \textit{\textbf{smil}ed}.
In order to investigate the importance of the latter, we further experiment with autoencoding of \textit{random strings} and find that for our experimental 
settings and non-templatic languages the performance gain is comparable to using corpus words.

\section{Model Description}
The log-likelihood for joint training on the tasks of MRI and autoencoding is:
\begin{align}
  {\cal L}({\boldsymbol \theta})\!=& \!\!\!\sum_{(f_s, f_t, t) \in {\cal T}} \!\!\!\!\log \ptheta\left(f_t \mid e(f_s, t) \right) \label{eq:ll} \\ 
  & + \text{ } \!\!\!\sum_{w \in {\cal W}} \text{ }\quad\!\!\!\! \log \ptheta (w \mid e(w)), \nonumber
\end{align}
$\cal T$ is the MRI training data, with each example consisting of a source form $f_s$, a target form $f_t$ and a target tag $t$. $\cal W$ denotes a
set of words in the language of the system.
The encoding function $e$ depends on $\boldsymbol\theta$.
The parameters ${\boldsymbol \theta}$ are shared across the two tasks, resulting in a share of information.
We obtain this by giving our model data from both sets at the same time, and marking each example with a
task-specific input symbol, cf. Figure \ref{fig:example}.
Following \cite{kann2016med}, we employ a neural encoder-decoder model.

\paragraph{Encoder. }
For the input of the encoder,
we adapt the format by \newcite{kann2016med}, but modify it to be able to handle unlabeled data:
Given the set of morphological subtags M each target tag is composed of (e.g., the tag \textit{1SgPresInd} contains the subtags
\textit{1}, \textit{Sg}, \textit{Pres} and \textit{Ind}), and the
alphabet $\Sigma$ of the language of application, our input is of the form ${\tt B}[\text{\tt \textbf{A}}/\text{M}^*]\Sigma ^*{\tt E}$, i.e.,
it consists of \textit{either} a sequence of subtags \textit{or} the symbol {\tt \textbf{A}} signaling that the input is not annotated and should be autoencoded,
and (in both cases) the character sequence of the input word. 
{\tt B} and {\tt E} are start and end symbols.
Each part of the input is represented by an embedding.

We then encode the input $x = x_1, x_2, \ldots, x_{T_x}$ using a bidirectional gated recurrent neural network (GRU) \cite{cho-al-emnlp14},
i.e., $\overrightarrow{h}_i = f\left(\overrightarrow{h}_{i-1},
x_i\right)$ and $\overleftarrow{h}_i = f\left(\overleftarrow{h}_{i+1},
x_i\right)$, with $f$ being the update function of the hidden layer\katha{Check this!}.
Forward and backward hidden states are concatenated to obtain the input $h_i$ for the decoder.

\paragraph{Decoder. }
The decoder is an attention-based GRU, defining a probability distribution over
strings in $\Sigma^*$:
\enote{kk}{Should this be $p(y \mid h_i)$ as this is ONLY the decoder?}
\begin{equation*}
  p(y \mid x) = \prod_{t=1}^{T_y} p(y_t \mid y_1, \ldots, y_{t-1}, s_t, c_t),
\end{equation*}
with $s_t$ being the decoder hidden state for time $t$
and $c_t$ being a context vector, calculated using the encoder hidden states together with attention weights.
A detailed description of the model can be found in \newcite{bahdanau2014neural}.

\section{Experiments}
\label{sec:experiment}
\begin{table*}[!htbp]
  \centering
    \setlength{\tabcolsep}{1.5pt} 
  \tiny
\begin{tabular}{ll||c c c c | c c c c | c c c c | c c c c | c c c c | c c c c | c c c c | c c c c }
    & & \multicolumn{4}{c|}{\textbf{ar}} & \multicolumn{4}{c|}{\textbf{fi}} &\multicolumn{4}{c|}{\textbf{ka}} &\multicolumn{4}{c|}{\textbf{de}} &\multicolumn{4}{c|}{\textbf{nv}} &\multicolumn{4}{c|}{\textbf{ru}} &\multicolumn{4}{c|}{\textbf{sp}} &\multicolumn{4}{c}{\textbf{tu}}\\
    & & \rotatebox{90}{SIG16} & \rotatebox{90}{SIG17} & \rotatebox{90}{MED} & \rotatebox{90}{Our} & \rotatebox{90}{SIG16} & \rotatebox{90}{SIG17} & \rotatebox{90}{MED} & \rotatebox{90}{Our} & \rotatebox{90}{SIG16} & \rotatebox{90}{SIG17} & \rotatebox{90}{MED} & \rotatebox{90}{Our} & \rotatebox{90}{SIG16} & \rotatebox{90}{SIG17} & \rotatebox{90}{MED} & \rotatebox{90}{Our} & \rotatebox{90}{SIG16} & \rotatebox{90}{SIG17} & \rotatebox{90}{MED} & \rotatebox{90}{Our} & \rotatebox{90}{SIG16} & \rotatebox{90}{SIG17} & \rotatebox{90}{MED} & \rotatebox{90}{Our} & \rotatebox{90}{SIG16} & \rotatebox{90}{SIG17} & \rotatebox{90}{MED} & \rotatebox{90}{Our} & \rotatebox{90}{SIG16} & \rotatebox{90}{SIG17} & \rotatebox{90}{MED} & \rotatebox{90}{Our} \\  \hline
    \multirow{2}{*}{\textbf{$\frac{1}{4}$}}  & acc & .188 & .094 & .716 & {\bf .722} & .293 & .325 & .809 & {\bf .854} & .814 & .831 & .910 & {\bf .912} & .721 & .687 & .882 & {\bf .888} & .317 & .403 & .706 & {\bf .711} & .641 & .638 & {\bf .825} & .824 & .558 & .539 & .939 & {\bf.942} & .181 & .129 & .904 & {\bf .910} \\
                                             & ED & 2.26 & 3.06 & 0.94 & {\bf 0.92} & 1.90 & 1.47 & 0.47 & {\bf 0.35} & 0.42 & 0.38 & {\bf 0.28} & 0.30 & 0.47 & 0.54 & 0.33 & {\bf 0.31} & 2.04 & 1.95 & 1.01 & {\bf 0.97}& 0.69 & 0.65 & {\bf 0.43} & {\bf 0.43} & 0.96 & 0.97 & {\bf 0.15} & {\bf 0.15} & 2.92 & 3.33 & 0.27 & {\bf 0.23} \\ \hline
    \multirow{2}{*}{\textbf{$\frac{1}{8}$}}  & acc & .104 & .063 & .600 & {\bf .640} & .207 & .227 & .687 & {\bf .732} & .798 & .791 & .883 & {\bf .894} & .618 & .593 & .851 &{\bf .873} & .247 & .350 & .516 & {\bf .619} & .516 & .523 & .766 & {\bf .772} & .441 & .409 & .896 & {\bf .916} & .120 & .080 & {\bf .846} & .832 \\
                                             & ED & 2.76 & 3.32 & 1.37 & {\bf 1.20} & 2.32 & 1.91 & 0.85 & {\bf 0.77} & 0.47 & 0.44 & 0.45 & {\bf 0.42} & 0.67 & 0.73 & 0.42 & {\bf 0.35} & 2.40 & 2.23 & 1.75 & {\bf 1.40} & 0.95 & 0.92 & {\bf 0.60} & {\bf 0.60} & 1.36 & 1.35 & 0.26 & {\bf 0.22} & 3.42 & 3.80 & {\bf 0.47} & 0.54 \\ \hline
    \multirow{2}{*}{\textbf{$\frac{1}{16}$}} & acc & .052 & .043 & .470 & {\bf .533} & .126 & .149 & .543 & {\bf .620} & .709 & .751 & .860 & {\bf .875} & .504 & .495 & .791 & {\bf .839} & .204 & .329 & .350 & {\bf .473} & .384 & .422 & .645 & {\bf .695} & .317 & .308 & .807 & {\bf .862} & .070 & .049 & .717 & {\bf .739} \\
                                             & ED & 3.36 & 3.53 & 1.80 & {\bf 1.59} & 2.84 & 2.34 & 1.33 & {\bf 1.16} & 0.62 & {\bf 0.50} & 0.58 & 0.52 & 0.90 & 0.94 & 0.60 & {\bf 0.45}& 2.71 & 2.41 & 2.63 & {\bf 2.05} & 1.23 & 1.17 & 0.94 & {\bf 0.82} & 1.80 & 1.70 & 0.47 & {\bf 0.36} & 3.81 & 4.09 & 0.99 & {\bf 0.94} \\ \hline
    \multirow{2}{*}{\textbf{$\frac{1}{32}$}} & acc & .028 & .027 & .263 & {\bf .381} & .073 & .088 & .314 & {\bf.402} & .595 & .648 & .818 & {\bf .852} & .384 & .386 & .661 & {\bf .722} & .174 & .303 & .174 & {\bf .369} & .249 & .293 & .406 & {\bf .502} & .196 & .245 & .657 & {\bf .756} & .044 & .028 & .524 & {\bf .571} \\
                                             & ED & 3.73 & 3.73 & 2.79 & {\bf 2.22} & 3.18 & 2.76 & 2.48 & {\bf 2.00} & 0.87 & 0.70 & 0.76 & {\bf 0.65} & 1.15 & 1.18 & 1.01 & {\bf 0.90} & 2.94 & {\bf 2.65} & 3.85 & 2.73 & 1.61 & 1.45 & 1.71 & {\bf 1.38} & 2.22 & 2.06 & 0.97 & {\bf 0.62} & 4.19 & 4.27 & 1.98 & {\bf 1.80} \\
  \end{tabular}
\caption{Accuracy (the higher the better) and edit distance (the lower the better) for our system and the three baselines on the official test set of task 3 of the SIGMORPHON 2016 shared task. Only the indicated amount (row labels) of
the original training data is used, emulating a low-resource setting. Best results for each language in bold.\label{results:exp1}}
\end{table*}
\paragraph{Dataset. }
We experiment on the task 3 dataset of the SIGMORPHON 2016 shared task on MRI \cite{cotterell-et-al-2016-shared}
and all standard languages provided:
Arabic, Finnish, Georgian, German, Navajo, Russian, Spanish and Turkish. 
German, Spanish and Russian are suffixing and
exhibit stem changes. Russian differs from the other two in that those stem changes are consonantal and not vocalic. Finnish and Turkish
are agglutinating, almost exclusively suffixing and have vowel harmony systems. 
Georgian uses both prefixiation and suffixiation. 
In contrast, Navajo mainly makes use of prefixes with consonant harmony among its sibilants.
Finally, Arabic is a templatic, non-concatenative language.

For each language, we further add randomly sampled words from the respective Wikipedia dumps. 
We exclude tokens that are not exclusively composed from characters of the language's alphabet,
e.g., digits, or do not appear at least 2 times in the corpus.
The exact amount of unlabaled data added is treated as a hyperparameter depending on the number of available annotated examples and optimized 
on the development set, cf. Section \ref{subsec:analysis}. 
Evaluation is done on the official shared task test set.

\paragraph{Training, hyperparameters and evaluation. } 
We mainly adopt the hyperparameters of \cite{kann2016med}.
Embeddings are 300-dimensional, the size of all hidden
layers is 100 and for training we use {\sc AdaDelta} \cite{abs-1212-5701} with a batch size of 20.
We train all models which use $\frac{1}{8}$ or more of the labeled data for 200 epochs, and models that see $\frac{1}{16}$ and $\frac{1}{32}$ of the original data for 400 and 800 epochs, respectively.
In all cases, we apply the last model for testing.

We evaluate using two metrics: accuracy and edit distance. Accuracy reports the percentage of completely correct solutions,
while the edit distance between the system's guess and the gold solution gives credit to systems that produce forms that
are close to the right form.

\paragraph{Baselines. }
We compare our system to three baselines\katha{Add the 2016 shared task system here as well}:
The first one is \textbf{MED}\footnote{http://cistern.cis.lmu.de/med/}, the winning system of the 2016 shared task. 
The network architecture is the same as in our system, 
but it is trained exclusively on 
labeled data. Thus, we expect it to suffer stronger from a lack of resources.

The second baseline is the official SIGMORPHON 2016 shared task baseline (\textbf{SIG16}) \cite{cotterell-et-al-2016-shared}, which is 
similar in spirit to the system described by \newcite{nicolai2015inflection}. 
The system treats the prediction of edit operations to be performed on the input string as a sequential decision-making problem,
greedily choosing each edit action given the previously chosen actions.
The selection of operations is made by an averaged perceptron,
using the binary features described in
\cite{cotterell-et-al-2016-shared}.\footnote{Note that our use of the
system differs from the official baseline in that we perform a direct form-to-form mapping. The shared task system predicts first form-to-lemma and then lemma-to-form.
However, we assume no lemmata to be given, and thus are unable to train such a system.}

Third, we compare to the baseline system of the CoNLL-SIGMORPHON 2017 shared task on universal morphological reinflection (\textbf{SIG17}) \cite{cotterell-conll-sigmorphon2017},
which is extremely suitable for low-resource settings.
It splits all source and target forms in the training set into prefix, middle part and suffix,
and uses those to find prefix or suffix substitution rules. 
Every evaluation example is searched for the longest contained prefix or suffix and the
rule belonging to the affix and given target tag is applied to obtain the output.

\paragraph{Results and discussion.}
As shown in Table \ref{results:exp1}, additionally training on unlabeled examples improves the performance of the encoder-decoder network for nearly all settings and languages,
especially for the very low-resource scenarios with $\frac{1}{16}$ and $\frac{1}{32}$ of the training data. The biggest increase in accuracy can be seen for 
Russian and Spanish, both in the $\frac{1}{32}$ setting, with $0.0963$ ($0.5023 - 0.4060$) and $0.0992$ ($0.7564 - 0.6572$), respectively.  
For the settings with bigger amounts of training data available, the unlabeled data does not change performance a lot. This was expected, as the model already gets 
enough information from the annotated data. However, semi-supervised training never \textit{hurts} performance, and can thus always be employed.
Overall, our semi-supervised training method shows to be a useful extension of the original system.

Furthermore, there is only one case---Georgian, $\frac{1}{16}$---where any of the SIGMORPHON baselines outperforms the neural methods. This clearly
shows the superiority of neural networks for the task and emphasizes the need to reduce the amount of labeled training data required for their training.

\section{Analyses}

\subsection{Amount of Unlabeled Data}
\label{subsec:analysis}
\begin{figure}
\centering
\begin{minipage}{\columnwidth}
  \centering
  \includegraphics[width=.85\columnwidth]{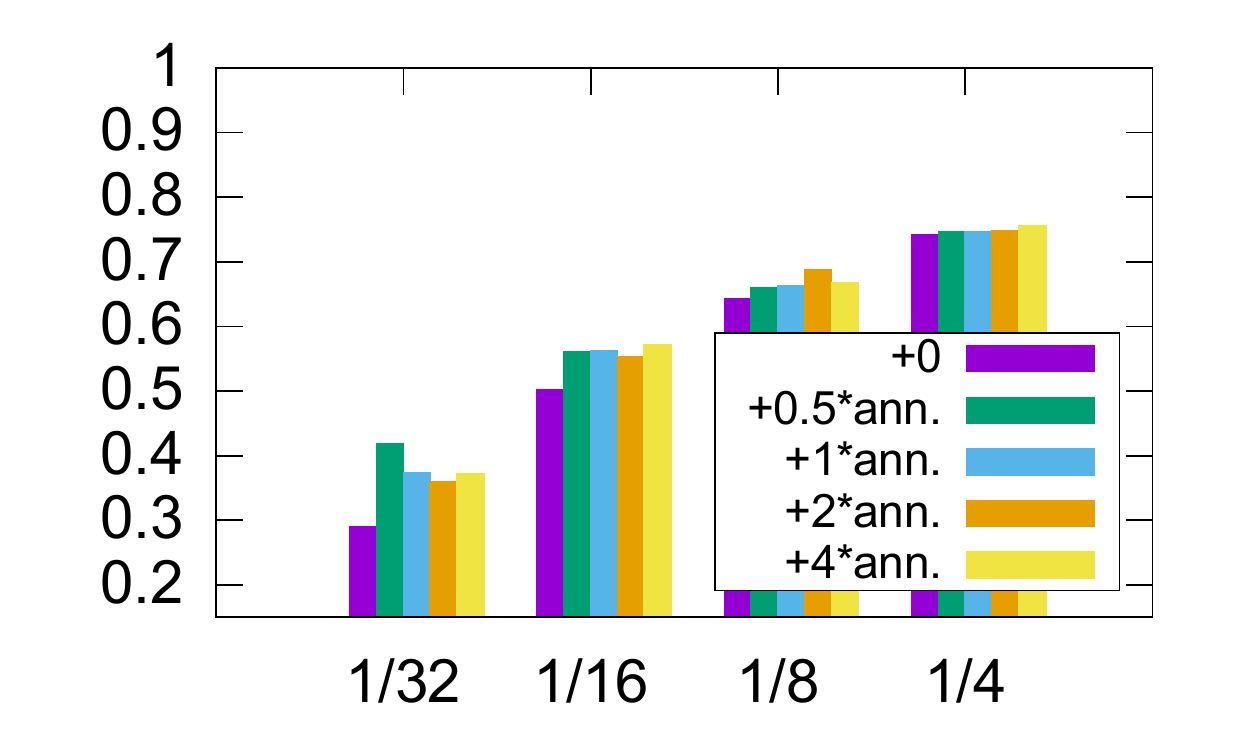}
  \caption*{Arabic}
  \label{fig:test1}
\end{minipage} \\
\begin{minipage}{\columnwidth}
  \centering
  \includegraphics[width=.85\columnwidth]{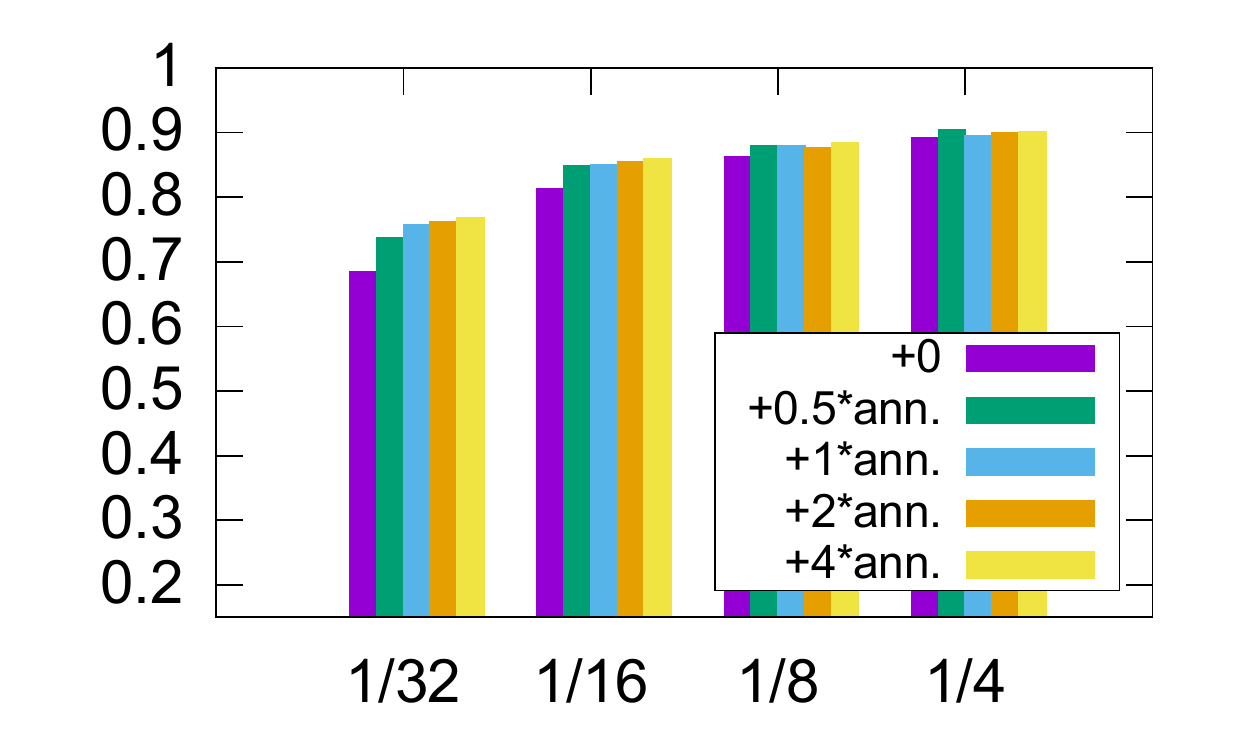}
  \caption*{German}
  \label{fig:test2}
\end{minipage}
\caption{Comparison of different amounts of unlabeled data, sorted by the amount of labeled training examples in portions of the original data.
Evaluated on the development set.} 
\label{fig:analysis}
\end{figure}
We now consider the amount of unlabeled examples as a function of the number of annotated examples.
Data and training regime are the same as in Section \ref{sec:experiment}. 
This analysis is performed on the development set and we report the highest accuracy obtained during training.

The resulting accuracies for Arabic and German can be seen in Figure \ref{fig:analysis}. The other languages behave similarly to German.
The loss of performance for reducing the training data varies a lot between languages, depending on how regular and thus "easy to learn" those are.
Concerning the amount of unlabeled examples, it seems that even though in single cases other ratios are slightly better, using 4 times more unlabeled examples
mostly obtains highest accuracy. Thus, a general rule could be that the more additional examples are used the better.
The only exception is Arabic in the $\frac{1}{32}$ setting,
where using half as many unlabeled as labeled examples obtains 
much better results. We explain this with the Semitic language being templatic. Since words in Arabic paradigms do not share a connected stem, we expect
that giving the model too much bias to copy might be harming performance in low-resource settings.
However, even for low-resource Arabic, using a ratio of 1:4 of labeled to unlabeled examples still yields a better performance than not using unlabeled examples at all.
Thus, we can conclude that if aiming for a language-independent setup, this is a good ratio.

\subsection{Autoencoding of Random Strings}
\label{subsec:combi}

We expect the network to benefit from a bias to copy strings. 
This suggests that \textit{any} random combination of characters from the language's alphabet could be
autoencoded in order to improve the performance in low-resource settings. 
To verify this, we train models on new datasets with $\frac{1}{32}$ of the labeled examples from task 3 of the SIGMORPHON 2016 shared task and the optimal number
of unlabeled examples for each language, cf. \S\ref{subsec:analysis}. However, the unlabeled examples
are now random strings of a length between 3 and 20. All models are trained as before.
Accuracies on the official test sets are shown
in Table \ref{table:copyExperiment}, and compared to (i) training without unlabeled examples and (ii) the data being enhanced 
by corpus words. 
\begin{table}
  \begin{adjustbox}{width=\columnwidth}
    \setlength{\tabcolsep}{3pt} 
  \centering
  \begin{tabular}{l ||c c c c c c c c  }
                      & \textbf{ar} & \textbf{fi} & \textbf{ka} & \textbf{de} & \textbf{nv} & \textbf{ru} & \textbf{es} & \textbf{tu} \\\hline
  MED                 & .2628 & .3144 & .8184 & .6608 & .1738 & .4060 & .6572 & .5238 \\
  MED+\textit{corpus} & \textbf{.3811} & \textbf{.4015} & .8523 & .7221 & \textbf{.3688} & \textbf{.5023} & .7564 & \textbf{.5713} \\
  MED+\textit{random} & .3064 & .3793 & \textbf{.8531} & \textbf{.7313} & .3250 & .4958 & \textbf{.7676} & .5706 \\
  \end{tabular}
  \end{adjustbox}
  \caption{Accuracies for
  MED (\newcite{kann2016med}), MED+\textit{corpus} and MED+\textit{random}. Descriptions in the text.\label{table:copyExperiment}}
\end{table}
Several aspects of the results are eye-catching. First, for Arabic, the gap to the performance with corpus words is the biggest,
showing that indeed the tendency of languages to copy the stem when inflecting is playing an important role. 
Second, for some languages the performance gains for corpus words and random words are comparable.
Third, the performance of random strings is closer to the performance of corpus words the higher the overall accuracy is. 
The additional unlabeled examples might be acting as regularizers in this case.

Overall, this experiment shows clearly that giving the model a bias to copy strings helps for inflection in non-templatic languages, and that
random strings can improve a network for MRI.

\section{Related Work}
\label{sec:related_work}

For the SIGMORPHON 2016 and the CoNLL-SIGMORPHON 2017 shared
tasks \cite{cotterell-et-al-2016-shared,cotterell-conll-sigmorphon2017}, 
multiple MRI systems were developed, e.g., \cite{nicolai-EtAl:2016:SIGMORPHON,taji-EtAl:2016:SIGMORPHON,kann2016med,aharoni2016improving,ostling:2016:SIGMORPHON,
uzh-sigmorphon2017}. 
Encoder-decoder neural networks \cite{cho-EtAl:2014:SSST-8,sutskever2014sequence,bahdanau2014neural} performed best, such that we extend them in this work.
Earlier work on paradigm completion included \cite{faruqui-EtAl:2016:N16-1,nicolai2015inflection,durrett2013supervised}.
Work directly tackling MRI was more rare, e.g., \cite{dreyer-eisner:2009:EMNLP}.
Our work relates to the line of research on minimally supervised and unsupervised methods for morphology, e.g., 
\newcite{creutz2007unsupervised} and \newcite{goldsmith2001unsupervised} presenting the unsupervised morphological segmentation systems Morfessor
and Linguistica, or \cite{dreyer-eisner:2011:EMNLP,poon-cherry-toutanova:2009:NAACLHLT09,snyder2008unsupervised}. 
However, none of those focused directly on MRI or on training neural networks for morphology.
The only case we know of where this \textit{was} done was work by \newcite{kann-cotterell-schutze:2017}. They leveraged morphologically annotated data in a closely related
high-resource language to reduce the need for labeled data in the target language. This works well for similar languages, but has the shortcoming to
require annotations in such a language to be at hand.
A similar approach was presented by \newcite{ha2016toward} for machine translation (MT).
Unlabeled corpora were used for semi-supervised training of models for MT, e.g., by
\newcite{cheng2016semi,vincent2010stacked,socher-EtAl:2011:EMNLP,ramachandran2016unsupervised}.
Those approaches differ from ours, due to a fundamental difference between the two tasks: For MRI, the source vocabulary and the target vocabulary are mostly the same.
This makes it intuitive for MRI to train the final model jointly on MRI and autoencoding.

\section{Conclusion}
\label{sec:conclusion}

We presented a way of semi-supervised training of a state-of-the-art model for low-resource MRI, using words from an unlabeled corpus. 
We found that the best ratio of labeled to unlabeled data depends of the morphological
typology of the language. Finally, we showed that autoencoding random strings also increases performance, for some languages as much as using corpus words.

\section*{Acknowledgments}
We would like to thank the anonymous reviewers for their insightful comments.
This work was supported by DFG (SCHU2246/10).

\bibliography{emnlp2017}
\bibliographystyle{emnlp_natbib}

\end{document}